\documentclass[10pt,twocolumn,letterpaper]{article}

\usepackage[pagenumbers]{iccv} 

\usepackage{amsfonts}

\newcommand{\editingTask}{Get-In-Video Editing}
\newcommand{\datasetName}{GetIn-1M\xspace}
\newcommand{\modelName}{GetInVideo\xspace}
\newcommand{\modelNameShort}{GIV\xspace}
\newcommand{\benchmarkName}{GetInBench\xspace}

\newcommand{\pipelineName}{Recognize-Track-Erase\xspace}

\definecolor{iccvblue}{rgb}{0.21,0.49,0.74}
\usepackage[pagebackref,breaklinks,colorlinks,allcolors=iccvblue]{hyperref}

\def\paperID{3539} 
\def\confName{ICCV}
\def\confYear{2025}

\title{Get In Video: 
Add Anything You Want to the Video}

\author{
    Shaobin Zhuang$^{1\spadesuit*}$\quad
    Zhipeng Huang$^{3\spadesuit*}$\quad
    Binxin Yang$^{2}$\quad
    Ying Zhang$^{2}$\quad
    Fangyikang Wang$^{6\spadesuit}$\quad\\
    Canmiao Fu$^{2}$\quad
    Chong Sun$^{2}$\quad
    Zheng-Jun Zha$^{3}$\quad
    Chen Li$^{2}$\quad
    Yali Wang$^{4,5\textsuperscript{$\dagger$}}$\quad
    \\
    \footnotesize{$^1$Shanghai Jiao Tong University
    \quad $^2$WeChat, Tencent Inc \quad} 
    \footnotesize{$^3$University of Science and Technology of China \quad} \\
    \footnotesize{$^4$Shenzhen Institute of Advanced Technology, Chinese Academy of Sciences  \quad} \\
    \footnotesize{$^5$Shanghai Artificial Intelligence Laboratory \quad $^6$Zhejiang University \quad} \\
    {\tt\small \url{https://zhuangshaobin.github.io/GetInVideo-project/}}
}

\begin{document}

\twocolumn[{%
\renewcommand\twocolumn[1][]{#1}%
\maketitle

\begin{center}
    \includegraphics[width=\linewidth]{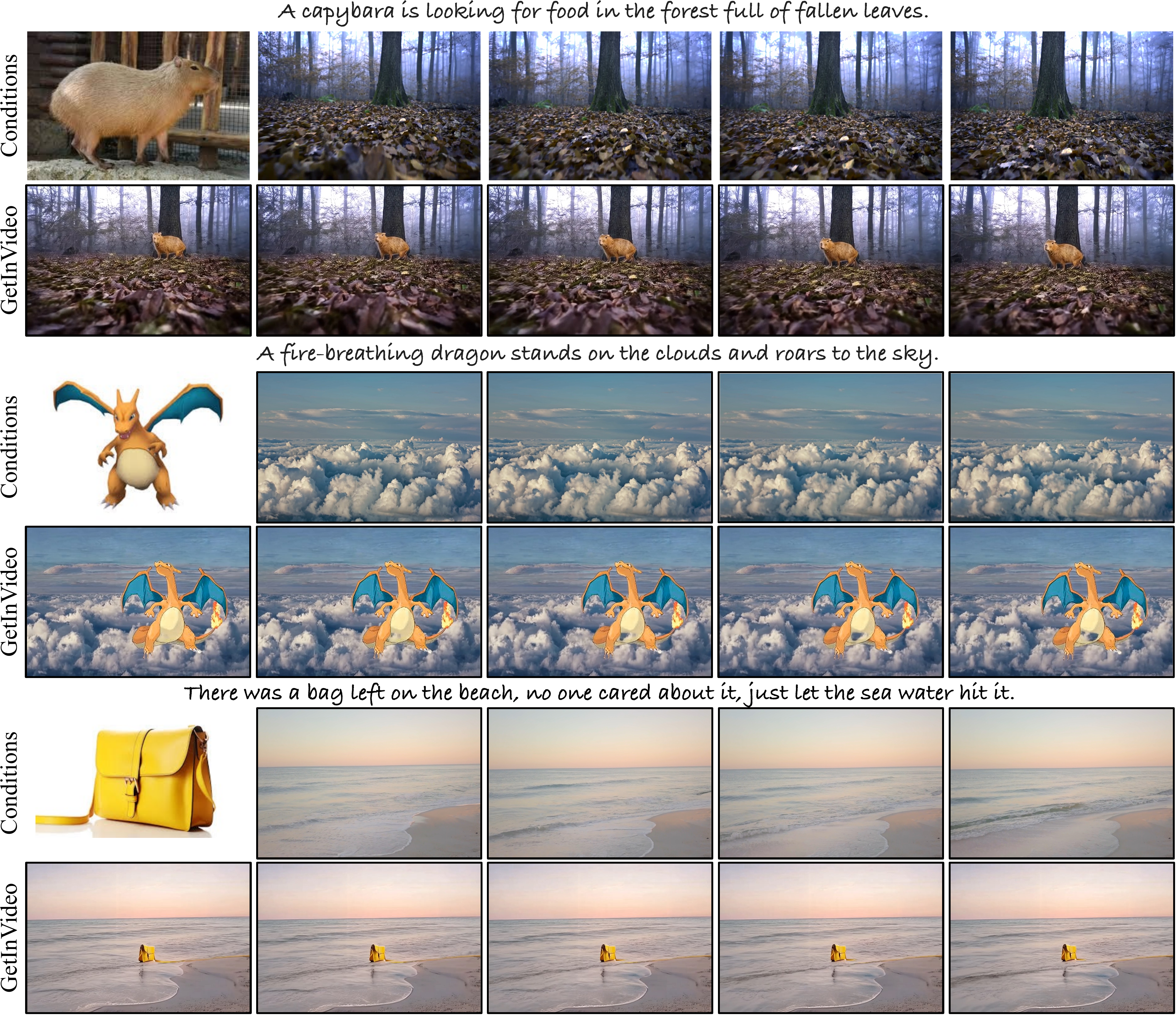}%
    \captionof{figure}{
    \textbf{Generated samples.} 
    Our \modelName model can accurately add instances from a reference image to a given video based on the prompt, while maintaining consistency between the instances and the environment and objects in the original video.
    }
    \label{fig:teaser}
    \vspace{6px}
\end{center}%
}]

\renewcommand{\thefootnote}{*}\footnotetext{Equal contribution \textsuperscript{$\dagger$} Corresponding author}
\renewcommand{\thefootnote}{$\spadesuit$}\footnotetext{Work done as interns at WeChat}

\begin{abstract}
Video editing increasingly demands the ability to incorporate specific real-world instances into existing footage, yet current approaches fundamentally fail to capture the unique visual characteristics of particular subjects and ensure natural instance/scene interactions. We formalize this overlooked yet critical editing paradigm as ``\editingTask", where users provide reference images to precisely specify visual elements they wish to incorporate into videos.
Addressing this task's dual challenges, severe training data scarcity and technical challenges in maintaining spatiotemporal coherence, we introduce three key contributions. 
First, we develop \datasetName dataset created through our automated \pipelineName pipeline, which sequentially performs video captioning, salient instance identification, object detection, temporal tracking, and instance removal to generate high-quality video editing pairs with comprehensive annotations (reference image, tracking mask, instance prompt). 
Second, we present \modelName, a novel end-to-end framework that leverages a diffusion transformer architecture with 3D full attention to process reference images, condition videos, and masks simultaneously, maintaining temporal coherence, preserving visual identity, and ensuring natural scene interactions when integrating reference objects into videos. 
Finally, we establish \benchmarkName, the first comprehensive benchmark for \editingTask\xspace
scenario, demonstrating our approach's superior performance through extensive evaluations. Our work enables accessible, high-quality incorporation of specific real-world subjects into videos, significantly advancing personalized video editing capabilities.
\end{abstract}
\section{Introduction}
\label{sec:intro}

\begin{figure*}[t]
    \centering
    \includegraphics[width=\textwidth]{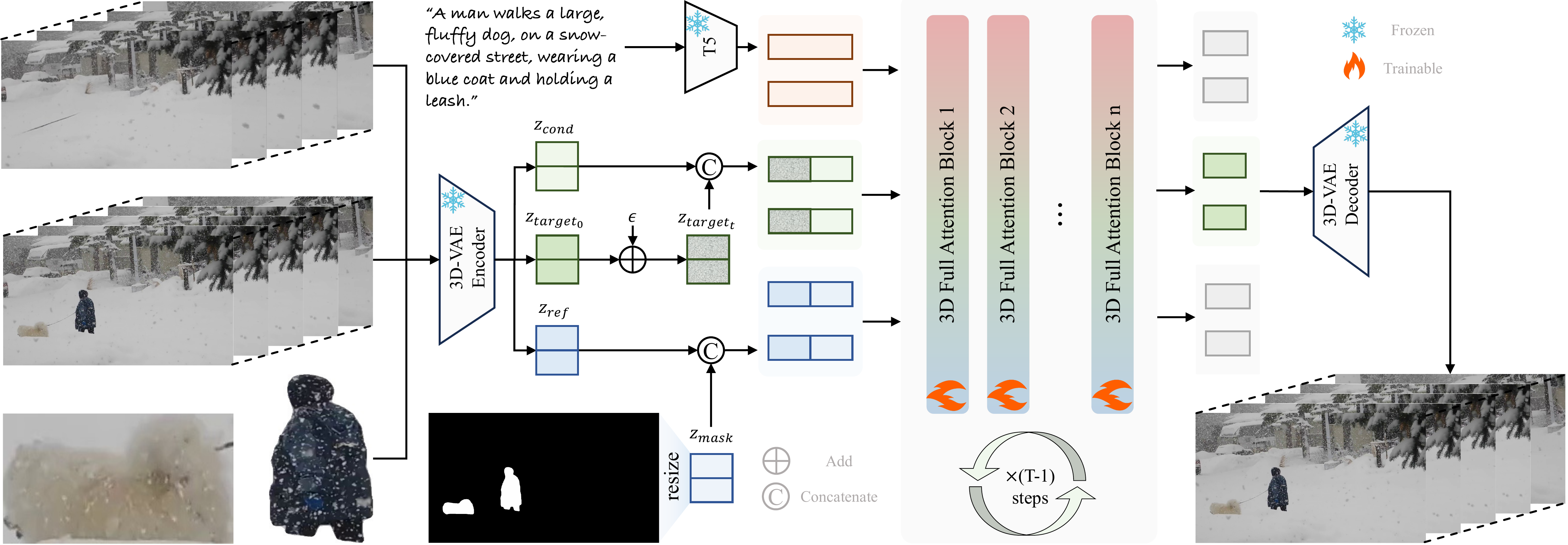}
    \caption{
    \textbf{The model architecture of \modelName}. 
    }
    \label{fig:model_arch}
\end{figure*}

Video editing has become increasingly important in various applications, including film production, social media content creation, and virtual reality experiences~\cite{sun2024diffusion,huang2024diffusion}. Traditional video editing approaches primarily focus on video clipping and material splicing, which not only require extensive manual effort and professional expertise, limiting accessibility for average users~\cite{rana2021deepremaster, leake2017computational}, but also offer limited functionality in terms of content manipulation and creative possibilities. 
Recent advancements in AI-driven editing tools have made significant progress in automating various aspects of video manipulation, ranging from text-guided editing~\cite{video-p2p, vid2vid-zero, fate-zero} to pose/point-guided editing~\cite{dragvideo,drag-a-video,follow-your-pose,dreampose,magic-animate}, motion guided style transfer~\cite{ground-a-video,magic-edit,ccedit,flowvid,anyv2v}, and video object removal~\cite{zhou2023propainter, yu2023deficiency}.



However, existing methods \cite{anyv2v, yang2023paint, brooks2022instructpix2pix} predominantly rely on text prompts or predefined operations, which have inherent limitations when users want to communicate specific visual details. This limitation becomes particularly evident when incorporating real-world instances (such as a particular person, pet, or object) into videos. Textual descriptions fundamentally fall short of conveying the unique visual characteristics that define these subjects~\cite{yang2023paint}, as language simply cannot capture the distinctive appearance features that make individuals recognizable.

We define this overlooked yet critical editing paradigm as ``\editingTask", where users provide reference images to precisely specify visual elements they wish to incorporate into videos. This task presents two significant challenges. First, there is a severe scarcity of training data that pairs reference images with appropriately edited videos. Second, successful implementation requires that inserted objects be coherently embedded within the video context, exhibiting natural motion and plausible interactions with existing scene elements. These requirements demand strong temporal consistency capabilities as the model must adapt the reference subject to varying poses, lighting conditions, and perspectives throughout the video sequence. To address these challenges:

we introduce \textbf{\datasetName} dataset, a million-scale dataset created through our automated \pipelineName pipeline. This comprehensive pipeline integrates multiple state-of-the-art components: NVILA \cite{liu2024nvila} for detailed video captioning, LLaMA-2 \cite{touvron2023llama} for identifying salient instances with significant motion and interaction, Ground-DINO \cite{liu2023grounding} for precise instance detection, SAM2 \cite{ravi2023sam2} for accurate temporal instance tracking, and ProPainter \cite{zhou2023propainter} for clean instance removal. The resulting training samples—\textit{\{prompt, reference image, tracking mask, condition video (object-removed), target video (original)\}}create an elegant supervised learning paradigm where the model learns to reinsert removed instances based on reference images, with original videos providing perfect ground truth supervision.

Based on our large-scale dataset, we develop \textbf{\modelName}, a novel end-to-end framework that leverages a diffusion transformer architecture with 3D full attention to seamlessly integrate reference objects into target videos. Our model simultaneously processes four key inputs: prompts describing instances in the video, reference images providing visual appearance, mask indicating position (optional), and condition videos with instances removed. Then \modelName outputs the edited video. Through our innovative latent space concatenation approach, all inputs inherently interact through the attention mechanism without requiring supplementary modules, enabling effective inheritance of pre-trained T2V parameters with only minimal adjustments to input layer dimensions. This architectural design ensures optimal preservation of pre-trained knowledge while achieving efficient multimodal integration, addressing the technical challenges of maintaining temporal coherence, preserving visual identity, and ensuring natural scene interactions.

To enable systematic evaluation, we introduce \textbf{\benchmarkName}, the first comprehensive benchmark specifically designed for reference-guided video object addition. This benchmark supports both qualitative and quantitative assessment through carefully curated test cases featuring both single-object and multi-object reference scenarios. Through extensive experimentation, \modelName demonstrates exceptional performance on this benchmark, outperforming existing approaches in visual quality and achieving significant quantitative advantages across multiple evaluation metrics.
\textit{In addition, we will open source the models, datasets, benchmarks, and codes mentioned above.}

Our principal contributions are as follows:
\begin{itemize}
\item We formalize \editingTask\xspace as a new editing scenario, addressing the fundamental limitations of previous approaches for incorporating specific real-world instances into videos.

\item We introduce \datasetName, a large-scale dataset specifically constructed for this task, containing 1 million video editing pairs with comprehensive textual annotations and reference images.

\item We develop \modelName, the first end-to-end model that effectively solves this challenging problem by maintaining temporal coherence, visual identity preservation, and natural scene interactions.

\item We establish \benchmarkName, the first comprehensive benchmark for this editing scenario, and demonstrate our approach's superior performance through extensive qualitative and quantitative evaluations.
\end{itemize}

\section{Related Work}
\label{sec:related_work}

\subsection{Video Generation}
Video generation witnesses remarkable progress in recent years. Prior to Sora~\cite{sora2024}, early works like Emu Video~\cite{girdhar2023emu} and Microcinema~\cite{wang2024microcinema} extend image-pretrained diffusion models to the video domain, demonstrating the potential of capturing temporal dynamics despite limitations in video length and quality. Following Sora's announcement, several closed-source commercial platforms including Runway Gen-3~\cite{runwaygen3}, Pika Labs~\cite{pikalabs}, and others achieve comparable capabilities, offering sophisticated video generation tools while enhancing temporal consistency and scene complexity handling. More recently, high-performance open-source models such as Hunyuan-Video~\cite{hunyuanvideo2024}, StepFun~\cite{wang2024stepfun}, and CogVideo~\cite{hong2022cogvideo} emerge, providing essential building blocks for complex video editing tasks like \editingTask\xspace by offering robust video modeling capabilities required for sophisticated temporal content manipulation.

\begin{figure*}[t]
    \centering
    \includegraphics[width=\textwidth]{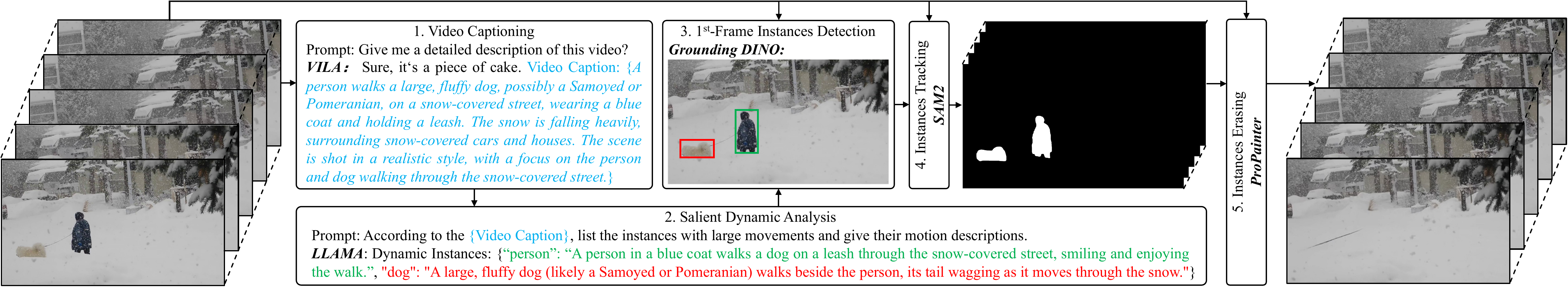}
    \caption{
    \textbf{The construction process of \datasetName.} 
    }
    \label{fig:data_pipe}
\end{figure*}

\subsection{Video Editing}
Video editing techniques evolve dramatically with AI integration, with approaches categorized by their control mechanisms and capabilities. Text-guided editing methods~\cite{video-p2p,vid2vid-zero,fate-zero} allow users to express editing intentions through textual prompts, offering intuitive interfaces but struggling to capture specific visual characteristics that cannot be adequately described through text. Structure-guided approaches~\cite{gen-1,magic-edit,ground-a-video,flowvid,wang2023videocomposer} provide more precise spatial control through conditioning signals like depth maps and optical flow, enhancing spatial precision at the cost of requiring technical expertise. Point-based interaction methods~\cite{dragvideo,drag-a-video} enable intuitive control by specifying control points, though they primarily focus on repositioning existing content rather than introducing new elements. Pose-guided techniques~\cite{follow-your-pose,dreampose,magic-animate} excel at human-centric editing but have limited applicability to other object types. 
Motion-guided approaches~\cite{ground-a-video,magic-edit,ccedit,flowvid,anyv2v} leverage motion patterns from the original video to guide content transformation, typically focusing on style transfer or appearance modification of existing elements rather than coherent integration of new instances. 
Video object removal techniques~\cite{zhou2023propainter,yu2023deficiency} complement insertion tasks by providing mechanisms to prepare videos for new content integration. 
Unlike previous approaches, our \editingTask\xspace framework specifically addresses the challenge of incorporating particular real-world instances into videos with temporal coherence, identity preservation, and natural scene interactions.


The concurrent work VideoAnyDoor \cite{tu2025videoanydoor} employs reference images for video generation, producing entirely new video content without relying on original video inputs. 
In comparison, our proposed \editingTask\xspace Editing framework addresses a more complex video editing challenge: seamlessly integrating reference objects into existing videos while maintaining natural interactions with original scene elements. This technical distinction is crucial, as our solution must simultaneously preserve the contextual integrity of the source video and establish plausible physical relationships between inserted objects and the existing environment. Our approach specifically targets practical editing scenarios where users require precise subject incorporation into pre-recorded videos. Notably, we enhance editing flexibility through two key innovations: (1) optional mask guidance for precise spatial control, and (2) multi-object insertion capability - features absent in VideoAnyDoor's video generation-oriented framework, which focuses on creating new content rather than modifying existing footage.




\section{Method}
\label{sec:method}

Building upon our analysis of existing video editing approaches and their limitations, we now present \modelName, the first end-to-end framework specifically designed for reference-guided video object insertion.
Our approach addresses the fundamental challenges: maintaining temporal coherence, preserving visual identity, and ensuring natural scene interactions. 
In this section, we first review the theoretical foundations of text-to-video diffusion models (Sec. \ref{subsec:t2v_diffusion}), then detail the architecture and training methodology of \modelName (Sec. \ref{subsec:method_model}), and finally introduce our \datasetName dataset and its data-pipeline (Sec. \ref{subsec:method_dataset}), which enables effective training for this novel editing task.

\subsection{Text-to-Video Diffusion Model}
\label{subsec:t2v_diffusion}
The diffusion model is designed to learn a data distribution by gradually denoising a normally-
distributed variable. 
It has been widely used for video generation.
Text-to-Video (T2V) diffusion model usually uses 3D-VAE encoder to compress the image or video $x$ into $z$ in latent space as $z_{0}=\text{Encoder}(x)$, 
$z_{0}\in \mathbb{R}^{f\times c \times h \times w}$, 
$x\in \mathbb{R}^{F\times 3 \times H \times W}$,
where $F$, $H$ and $W$ refer to the number of frames, height and width of image or video.
For a single frame $i$ in video $x$, we denote it as $x^k \in \mathbb{R}^{3 \times H \times W}$, where $k \in \{0, 1, ..., F-1\}$.
$f$, $c$, $h$ and $w$ represent the shape of feature after 3D-VAE compression.
In \modelName, $f=\lceil \frac{F-1}{4} \rceil+1$, $c=16$, $h=\frac{H}{8}$ and $w=\frac{W}{8}$.
In the forward diffusion process, 
one should add gaussian noise $\epsilon \sim \mathcal{N} \left( 0, I \right)$ on the input $z_{0}$,
in order to generate the noisy input $z_{t}$ at each timestep,
\begin{equation}
z_{t}=\sqrt{\overline{\alpha}_{t}}z_{0}+\sqrt{1-\overline{\alpha}_{t}}\epsilon
\label{eq:zt}
\end{equation}
where $t = 1, 2, \cdots , T$, 
and $T$ is the total number of timesteps in the forward process.
$\overline{\alpha}_{t}$ is a parameter related to $t$.
When $t$ approaches $T$, 
$\overline{\alpha}_{t}$ approaches $0$.
The training goal is to minimize the loss for denoising,
\begin{equation}
\mathcal{L}=\mathbf{E}_{z_{0},c,\epsilon,t} \left[ \|\epsilon - \epsilon_{\Theta} \left(z_{t},t,c \right) \|^{2}_{2} \right],~~ t \in [1,~T]
\label{eq:diff_loss}
\end{equation}
where $\epsilon_{\Theta}$ is the output of neural network with model parameters $\Theta$, and $c$ indicates the additional condition.

\subsection{\modelName}
\label{subsec:method_model}
As shown in Fig. \ref{fig:model_arch}, our model \modelName (\modelNameShort) simultaneously processes five key inputs: (1) textual prompt $p_i$ describing the instance, (2) reference image $x_{\text{ref}}$ providing visual appearance, (3) temporal mask sequence $m^0$ indicating object position in the first frame (optional), (4) condition video $x_{\text{cond}}$ with the instance removed.
And the target video $x_{\text{target}}$ serving as ground truth during training.
In order to be able to input the unedited original video as  condition video for our \modelName, 
we expand the number of input channels $c$ ($c=16$ in \modelName) of the original input layer of the model to $2c$,
and used zero initialization for the parameters in the expanded channels to ensure that the input distribution of the T2V model would not be affected.
During training,
we input the condition video $x_{cond}$ and target video $x_{target}$ into 3D-VAE and get the compression latent $z_{cond}$ and $z_{target_0}$.
Following eq. \ref{eq:zt}, we add gaussian noise $\epsilon \sim \mathcal{N} \left( 0, I \right)$ on $z_{target_0}$,
\begin{equation}
z_{target_t}=\sqrt{\overline{\alpha}_{t}}z_{target_0}+\sqrt{1-\overline{\alpha}_{t}}\epsilon
\label{eq:ztarget}
\end{equation}
Then we concat $z_{cond}$ and $z_{target_t}$ in the feature dimension to get $z_{video_t}$,
\begin{equation}
z_{video_t}=\texttt{concat([}z_{target_t} \texttt{,} z_{cond} \texttt{],dim=1)}
\label{eq:zvideo}
\end{equation}
which $z_{video_t}\in R^{f\times 2c \times h \times w}$.
Likewise, 
we send all the reference pictures $x_{ref}=[x_{ref_{0}}, x_{ref_{1}}, ...,x_{ref_{n-1}}]$ into 3D-VAE to obtain the latent 
$z_{ref}\in R^{n \times c \times h \times w}$,
where $n$ refers to the number of objects used for reference video generation.
And we perform spatial downsampling and channel dimension repeat operations on the mask $m\in R^{1\times 1\times H \times W}$  in sequence,
\begin{equation}
z_{mask}=\texttt{resize(}m\texttt{,(}w\texttt{,} h\texttt{))}\texttt{.repeat(}n\texttt{,}c\texttt{,1,1)}
\label{eq:zmask}
\end{equation}
which $z_{mask}\in R^{n\times c\times h\times w}$.
Then we concat $z_{ref}$ and $z_{mask}$ in channel dimension to get $z_{image}$,
\begin{equation}
z_{image}=\texttt{concat([}z_{ref} \texttt{,} z_{mask} \texttt{],dim=1)}
\label{eq:zimage}
\end{equation}
Next, we concat the latent of $z_{image}$ and $z_{video_t}$ in the frame dimension to get the model input $z_{input_t}$,
\begin{equation}
z_{input_t} = \texttt{concat([} z_{ref}\texttt{,} z_{video_t} \texttt{],dim=0)}
\label{eq:zt_}
\end{equation}
which $z_{input_t}\in R^{\left( n+f \right) \times 2c \times h \times w}$.
This means that we use the first $n$ frames of the input video as reference pictures to participate in the generation.

Following previous works,
\modelName inherits the most advanced diffusion transformer architecture with 3D full attention.
This demonstrates that in the \modelName framework, 
all inputs and conditional information inherently interact through the attention mechanism without requiring supplementary modules, 
which enables effective inheritance of the original T2V pre-trained parameters with only minimal adjustments to the input layer's channel dimensions.
The architectural integrity ensures optimal preservation of pre-trained knowledge while achieving efficient multimodal integration.

\subsection{\datasetName Dataset}
\label{subsec:method_dataset}

To address the data scarcity challenge for \editingTask\xspace, we introduce \datasetName, a large-scale dataset specifically constructed for this novel task. The dataset contains 1 million video editing pairs with comprehensive annotations, created through our automated \pipelineName pipeline. We source raw video data $x \in \mathbb{R}^{F \times 3 \times H \times W}$ from diverse public datasets including OpenVid~\cite{wu2023openvid}, YouTube-8M~\cite{abu2016youtube}, and PexelsVideos~\cite{pexelsvideos}, ensuring variety in content, style, and scene complexity. Our data generation pipeline consists of four key stages:

\paragraph{Instance-aware Video Captioning.}
We first employ NVILA \cite{liu2024nvila}, a state-of-the-art vision-language model, to generate detailed captions for each video $x$. These captions provide rich contextual descriptions of the video content, including objects, actions, and scene elements. 

To extract specific instances and their descriptions, we process these captions using LLaMA-2 \cite{touvron2023llama} to identify salient objects and generate structured instance-description pairs $\{i, p_i\}$. We specifically utilize LLaMA-2 for this rewriting task because it can identify prominent instances with significant motion, interaction, or functional interaction with the video, rather than static background elements or insignificant objects. This approach aligns with user preferences for editing meaningful foreground elements and produces higher-quality, more learnable training data by focusing on instances that exhibit clear temporal dynamics and scene interactions. In this context, $i$ represents the instance identifier (e.g., ``person") and $p_i$ is the corresponding detailed prompt (e.g., ``A person in a blue coat walks a dog on a leash through the snow-covered street, smiling and enjoying
the walk").

\paragraph{Instance Detection.}
For each identified instance $i$, we extract the first frame of the video $x^0$ and use Ground-DINO \cite{liu2023grounding} to detect the bounding box $b_i$ of the instance:
\begin{equation}
b_i = \text{GroundDINO}(x^0, i)
\end{equation}
where $b_i$ represents the coordinates of the bounding box.

\paragraph{Temporal Instance Tracking.}
Using the detected bounding box $b_i$ as initialization, we employ SAM2 \cite{ravi2023sam2} to generate a precise segmentation mask for the instance in the first frame. We then track this instance throughout the entire video sequence to obtain a temporal segmentation mask $m_i \in \mathbb{R}^{F \times 1 \times H \times W}$:
\begin{equation}
m_i = \text{SAM2}(x, b_i)
\end{equation}
where $x \in \mathbb{R}^{F \times 3 \times H \times W}$ is the original video, and $m_i \in \mathbb{R}^{F \times 1 \times H \times W}$ is the binary mask sequence tracking the instance across all $F$ frames.

\paragraph{Instance Removal.}
Finally, we use ProPainter \cite{zhou2023propainter}, a state-of-the-art video inpainting model, to remove the tracked instance from the original video:
\begin{equation}
x_{\text{cond}} = \text{ProPainter}(x, m_i)
\end{equation}
where $x_{\text{cond}} \in \mathbb{R}^{F \times 3 \times H \times W}$ is the condition video with the instance removed. While ProPainter relies on optical flow for inter-frame information propagation and may produce minor artifacts in areas with subtle motion, these imperfections do not affect our model's performance since $x_{\text{cond}}$ serves as input during training while the original video $x$ provides the ground truth supervision.

For each processed video, we extract the first clear frame containing the instance as the reference image $x_{\text{ref}} \in \mathbb{R}^{3 \times H \times W}$. This results in comprehensive training samples:
\begin{equation}
\{p_i, x_{\text{ref}}, m_i, x_{\text{cond}}, x_{\text{target}}\}
\end{equation}
where $x_{\text{target}} = x$ is the original unmodified video. \textbf{It is crucial to note that in our training process, the original video $x$ serves directly as the target video $x_{\text{target}}$, and the condition video $x_{\text{cond}}$ is the video with the instance removed}. This creates an elegant supervised learning paradigm: our model learns to reinsert the removed instance back into the video based on the reference image, with the original video providing the perfect ground truth for supervision.

The \datasetName dataset enables effective training of our \modelName framework by providing diverse, high-quality examples of \editingTask\xspace scenarios. The automated pipeline ensures consistency and scalability, allowing us to generate sufficient data to train our model for this challenging task.

\section{Experiments}
\label{sec:exp}

\begin{figure*}[t]
    \centering
    \includegraphics[width=\textwidth]{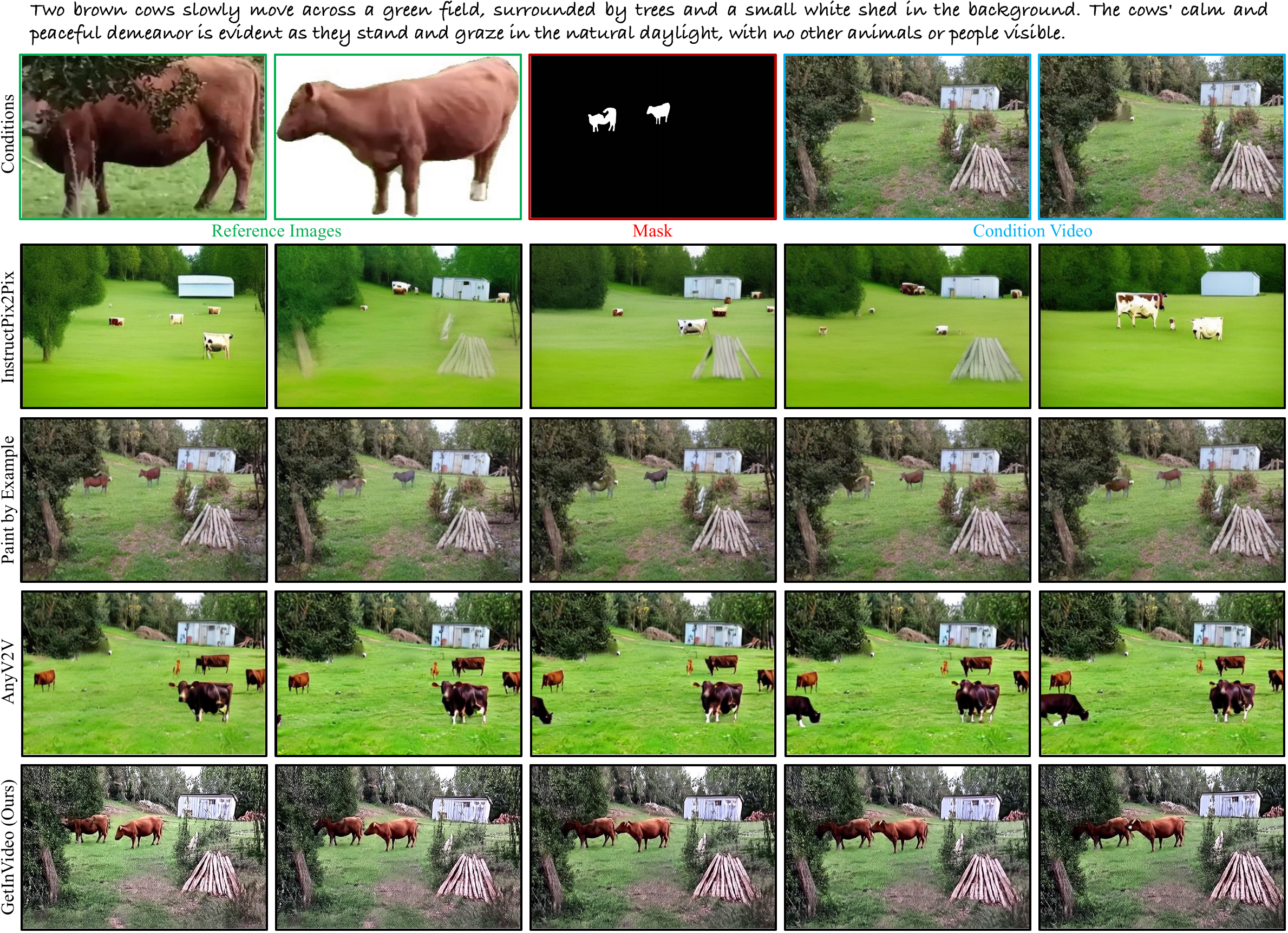}
    \caption{
    \textbf{Visual comparison with the SOTA model.} 
    }
    \label{fig:sota_compare}
\end{figure*}

\subsection{\benchmarkName}
\label{subsec:gib}

We introduce \benchmarkName, a benchmark specifically designed for reference image-based video object addition that contains 100 test cases constructed following the methodology of the \datasetName dataset. Each test case consists of five essential components: condition video, reference images, prompt, mask, and target video.
To evaluate video generation quality comprehensively, our benchmark employs five complementary metrics:
(1) Fréchet Inception Distance (FID) \cite{heusel2017gans}, which uses CLIP-ViT-Base-32 \cite{radford2021learning} features to measure the distribution difference between generated and target video frames;
(2) Fréchet Video Distance (FVD) \cite{unterthiner2019accurate}, which employs I3D \cite{Carreira2017QuoVA} features to assess temporal coherence and distribution similarity across complete video sequences;
(3) Image-to-Image CLIP Score (CLIP-I) \cite{Ye2023IPAdapterTC}, which evaluates the semantic similarity between generated and target videos at the conceptual level;
(4) Image-to-Image DINO Score (DINO-I), which captures fine-grained visual similarities through DINOv2-Base \cite{Oquab2023DINOv2LR} feature analysis; and
(5) Text-to-Image CLIP Score (CLIP-T) \cite{Wu2021GODIVAGO}, which measures the semantic alignment between generated videos and their corresponding text prompts.
This evaluation framework integrates spatial-temporal metrics (FID/FVD) with semantic consistency measures (CLIP-I/CLIP-T) and detailed visual correspondence (DINO-I), allowing for thorough assessment of both static visual quality and dynamic temporal characteristics.

\subsection{Implementation Details}
\label{subsec:imple}

\textbf{Training.}
We utilize CogVideoX-T2V 2B \cite{yang2024cogvideox} as our text-to-video (T2V) foundation model. Following the methodology described in Sec. \ref{subsec:method_model}, we expand the input layer channels from 16 to 32, with zero-initialization for the newly added parameters. Our experiments leverage the \datasetName dataset to train models at 480$\times$720 resolution with 25-frame.
To enhance temporal coherence, we maintain a consistent frame sampling interval of 1 during both training and inference. For data augmentation and robustness, we randomly drop 20\% of prompts, 20\% of reference images, and 50\% of masks during training. 

Optimization is conducted using AdamW \cite{Loshchilov2017DecoupledWD} with hyperparameters $\beta_{1}=0.9$, $\beta_{2}=0.95$, $\beta_{3}=0.98$, $\varepsilon=10^{-8}$, and weight decay of 1e-4. We employ a fixed learning rate of 5e-6 for training stability. The full-parameter fine-tuning process utilizes mixed-precision training (bfloat16) and DeepSpeed \cite{Rasley2020DeepSpeedSO} ZeRO-2 optimization, executed on 8$\times$A100 GPUs for 50,000 steps.

\noindent\textbf{Inference.}
During inference, we consistently employ DDIM \cite{Song2020DenoisingDI} sampling with 50 denoising steps. For standard T2V models, we apply classifier-free guidance \cite{Ho2022ClassifierFreeDG} throughout the generation process, formulated as:
\begin{equation}
\begin{split}
\widetilde{\epsilon}_{\Theta}(z_t,t,c_{\text{t}})&= \epsilon_{\Theta}(z_t,t,\varnothing) \\
&+s_1(\epsilon_{\Theta}(z_t,t,c_{\text{t}})-\epsilon_{\Theta}(z_t,t,\varnothing))
\label{eq:cfg}
\end{split}
\end{equation}
where $c_{\text{t}}$ represents the prompt condition and $s_1$ is the scaling parameter controlling the influence of the prompt.

In our \modelNameShort, the conditional inputs extend beyond textual prompts to include reference images and their corresponding masks, collectively forming the image-based condition $c_{\text{i}}$. Consequently, we extend the classifier-free guidance sampling procedure as:
\begin{equation}
\begin{split}
\widetilde{\epsilon}_{\Theta}(z_t,t,c_{\text{t}},c_{\text{i}})&= \epsilon_{\Theta}(z_t,t,\varnothing,\varnothing) \\
&+s_1(\epsilon_{\Theta}(z_t,t,c_{\text{t}},\varnothing)-\epsilon_{\Theta}(z_t,t,\varnothing,\varnothing)) \\
&+s_2(\epsilon_{\Theta}(z_t,t,c_{\text{t}},c_{\text{i}})-\epsilon_{\Theta}(z_t,t,c_{\text{t}},\varnothing))
\label{eq:icfg}
\end{split}
\end{equation}
where $c_{\text{i}}$ denotes the image condition and $s_2$ controls the influence of the image condition. For optimal results, we set $s_1=6$ and $s_2=1.5$ during inference.

\subsection{Comparison with state-of-the-art}
\label{subsec:sota}

\begin{table}[t]
    \centering 
    \setlength\tabcolsep{3pt}
    \resizebox{1.0\linewidth}{!}{%
    \begin{tabular}{l c c c c c}
        \toprule
        \textbf{Method} &  \textbf{FID ($\downarrow$)} &  \textbf{FVD ($\downarrow$)}  & \textbf{CLIP-I ($\uparrow$)} & \textbf{DINO-I ($\uparrow$)} & \textbf{CLIP-T ($\uparrow$)}  \\
        \midrule
        InstructPix2Pix \cite{brooks2022instructpix2pix} & 27.48 & 1235.49 & 0.7378 & 0.7833 & 0.3071 
        \\
        Paint by Example \cite{yang2023paint} & \textbf{14.83} & 918.70 & 0.8401 & 0.8792 & 0.2917 
        \\
        AnyV2V \cite{anyv2v} & 34.01 & 660.66 & 0.7442 & 0.7738 & 0.2887 
        \\
        \midrule
        \textbf{GIV (Ours)} & 15.12 & \textbf{435.22} & \textbf{0.8685} & \textbf{0.9267} & \textbf{0.3145} 
        \\
        \bottomrule
    \end{tabular}
    }
    \caption{\textbf{\benchmarkName SOTA Comparison.}}
    \vspace{-1.2em}
    \label{tab:gib_sota}
\end{table}

\textbf{Quantitative comparison.}
Tab. \ref{tab:gib_sota} presents a comparative analysis between our \modelNameShort framework and three representative approaches: the established end-to-end image editing model InstructPix2Pix \cite{brooks2022instructpix2pix}, the reference-based image editing model Paint by Example \cite{yang2023paint}, and the recent video editing method AnyV2V \cite{anyv2v}. Evaluation on the \benchmarkName benchmark reveals that \modelNameShort achieves superior performance across multiple key metrics. These results establish \modelNameShort as the first end-to-end video editing framework capable of executing precise reference image-guided modifications while maintaining temporal consistency and content fidelity.

\noindent\textbf{Qualitative comparison.}
As illustrated in Fig. \ref{fig:sota_compare}, our \modelNameShort framework demonstrates exceptional performance in detail preservation and object insertion. 
Unlike previous methods, \modelNameShort successfully maintains high fidelity to the original video content while seamlessly integrating reference objects into specific mask regions with pixel-level precision.
More results can be found in the supplementary material.

\begin{figure*}[t]
    \centering
    \includegraphics[width=\textwidth]{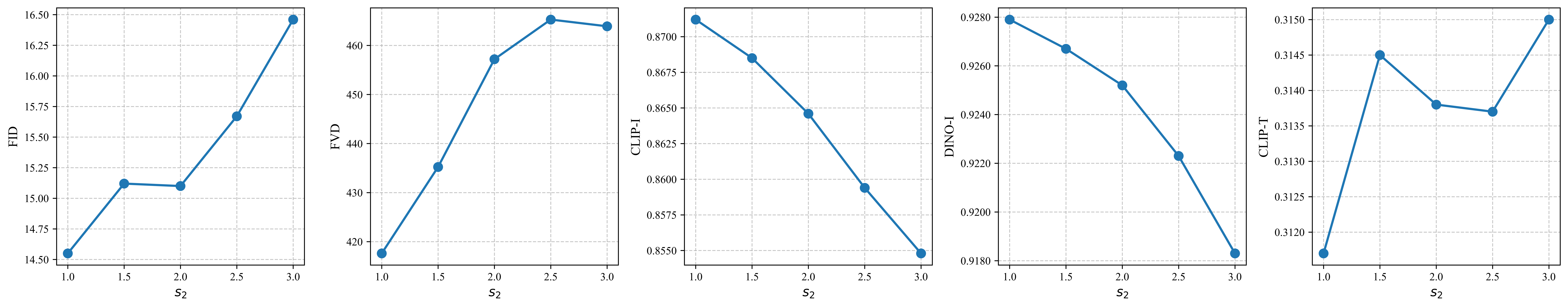}
    \vspace{-0.6cm}
    \caption{
    \textbf{The impact of classifier-free guidance on image condition.} 
    }
    \vspace{-0.2cm}
    \label{fig:cfg}
\end{figure*}

\begin{figure*}[t]
    \centering
    \includegraphics[width=\textwidth]{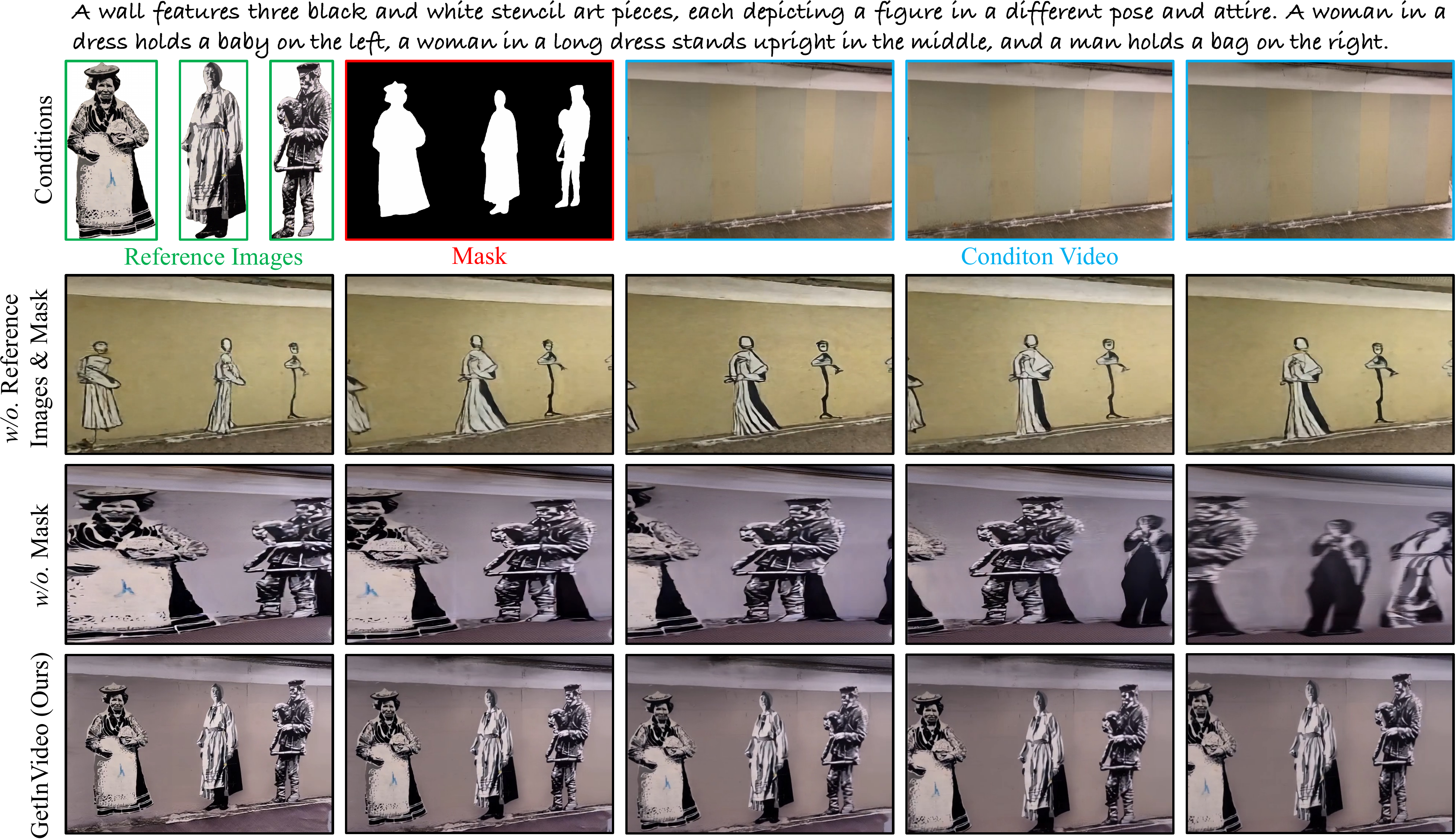}
    \caption{
    \textbf{Visual comparison with different input conditions.} 
    }
    \label{fig:abla_compare}
\end{figure*}

\subsection{Ablation Study}
\label{subsec:abla}

\textbf{Input conditions.}
As illustrated in Fig. \ref{fig:abla_compare}, 
we conducted ablation studies on different conditional inputs. 
When employing only textual prompts to edit the original video, 
the system demonstrated limited precision in controlling content generation within the synthesized frames. 
The incorporation of reference images as additional conditional inputs significantly improved instance-level accuracy, 
though spatial positioning remained suboptimal. 
Finally, 
through the integration of mask conditions, 
our \modelNameShort model achieved enhanced compliance with operational instructions, 
successfully integrating target instances from reference images into the original video while maintaining spatial coherence.

\noindent\textbf{Classifier-free guidance scale of image condition.}
Following previous work \cite{yang2024cogvideox},
We set the classifier-free guidance scale of text condition to 6 ($s_1=6$),
However, for our newly introduced image condition consisting of reference images and mask images, its classifier-free guidance scale is still worth exploring.
We perform ablation experiments on our \benchmarkName.
As shown in Fig. \ref{fig:cfg}, the effect is best when the classifier-free guidance scale of image condition is 1.5 ($s_2=1.5$).

\section{Conclusion}
\label{sec:conclusion}


We propose \editingTask\xspace Editing, an innovative paradigm that enables video editing through visual references instead of textual descriptions. This approach addresses three key challenges in video object insertion: (1) overcoming the inherent ambiguity of language in conveying visual details, (2) resolving training data scarcity, and (3) maintaining spatiotemporal consistency in video sequences. Our solution comprises three core components:
Firstly, we developed \datasetName - a million-sample dataset generated through an automated \pipelineName pipeline. This system produces precise annotation masks by detecting target objects across video frames, tracking their motion trajectories, and removing background elements.
Secondly, we designed \modelName, an end-to-end framework employing a diffusion transformer with 3D full-attention mechanisms. This architecture achieves three critical objectives: temporal consistency across frames, faithful preservation of object identity, and natural interaction with scene elements through adaptive lighting and perspective matching.
Thirdly, we established \benchmarkName, the first standardized evaluation benchmark for reference-guided video editing. Comprehensive experiments demonstrate our method's superior performance in visual quality and operational efficiency compared to existing approaches.

While significantly lowering the technical barrier for personalized video editing, current limitations in processing long video sequences warrant further investigation. We anticipate this work will guide future research directions in intuitive video editing tools, ultimately expanding creative possibilities in digital content creation.

{
    \small
    \bibliographystyle{ieeenat_fullname}
    \bibliography{main}

\begin{thebibliography}{51}
\providecommand{\natexlab}[1]{#1}
\providecommand{\url}[1]{\texttt{#1}}
\expandafter\ifx\csname urlstyle\endcsname\relax
  \providecommand{\doi}[1]{doi: #1}\else
  \providecommand{\doi}{doi: \begingroup \urlstyle{rm}\Url}\fi

\bibitem[Abu-El-Haija et~al.(2016)Abu-El-Haija, Kothari, Lee, Natsev, Toderici, Varadarajan, and Vijayanarasimhan]{abu2016youtube}
Sami Abu-El-Haija, Nisarg Kothari, Joonseok Lee, Paul Natsev, George Toderici, Balakrishnan Varadarajan, and Sudheendra Vijayanarasimhan.
\newblock Youtube-8m: A large-scale video classification benchmark.
\newblock In \emph{arXiv preprint arXiv:1609.08675}, 2016.

\bibitem[AI(2024)]{runwaygen3}
Runway AI.
\newblock Runway gen-3: Text-to-video generation.
\newblock \url{https://runwayml.com/}, 2024.
\newblock Accessed: 2024-05-01.

\bibitem[Brooks et~al.(2022)Brooks, Holynski, and Efros]{brooks2022instructpix2pix}
Tim Brooks, Aleksander Holynski, and Alexei~A Efros.
\newblock Instructpix2pix: Learning to follow image editing instructions.
\newblock \emph{arXiv preprint arXiv:2211.09800}, 2022.

\bibitem[Carreira and Zisserman(2017)]{Carreira2017QuoVA}
Jo{\~a}o Carreira and Andrew Zisserman.
\newblock Quo vadis, action recognition? a new model and the kinetics dataset.
\newblock \emph{2017 IEEE Conference on Computer Vision and Pattern Recognition (CVPR)}, pages 4724--4733, 2017.

\bibitem[Ceylan et~al.(2023)Ceylan, Huang, and Mitra]{video-p2p}
Duygu Ceylan, Chun-Hao~P Huang, and Niloy~J Mitra.
\newblock Video-p2p: Video editing with cross-attention control.
\newblock \emph{arXiv preprint arXiv:2303.04761}, 2023.

\bibitem[Chai et~al.(2023)Chai, Geng, Chen, Ren, Shan, Qiao, and Guo]{vid2vid-zero}
Wen Chai, Xian Geng, Zhaoxi Chen, Zhangyang Ren, Ying Shan, Yu Qiao, and Xiaohui Guo.
\newblock Vid2vid-zero: Zero-shot video editing using still image diffusion models.
\newblock \emph{arXiv preprint arXiv:2303.17599}, 2023.

\bibitem[Corran(2023)]{pexelsvideos}
Corran.
\newblock Pexelsvideos: A high-quality video dataset.
\newblock \url{https://huggingface.co/datasets/Corran/pexelvideos}, 2023.
\newblock Accessed: 2024-05-01.

\bibitem[Esser et~al.(2023)Esser, Kulal, Blattmann, Menick, Aghajanyan, Simonyan, and Rombach]{gen-1}
Patrick Esser, Sumith Kulal, Andreas Blattmann, Jacob Menick, Armen Aghajanyan, Karen Simonyan, and Robin Rombach.
\newblock Gen-1: Building a different kind of image generator.
\newblock \emph{arXiv preprint arXiv:2312.14148}, 2023.

\bibitem[Girdhar et~al.(2023)Girdhar, El-Nouby, Liu, Singh, Alwala, Mahajan, and Joulin]{girdhar2023emu}
Rohit Girdhar, Alaaeldin El-Nouby, Zhuang Liu, Mannat Singh, Kalyan~Vasudev Alwala, Dhruv Mahajan, and Armand Joulin.
\newblock Emu video: Factorizing text-to-video generation by explicit image conditioning.
\newblock \emph{arXiv preprint arXiv:2311.10709}, 2023.

\bibitem[Heusel et~al.(2017)Heusel, Ramsauer, Unterthiner, Nessler, and Hochreiter]{heusel2017gans}
Martin Heusel, Hubert Ramsauer, Thomas Unterthiner, Bernhard Nessler, and Sepp Hochreiter.
\newblock Gans trained by a two time-scale update rule converge to a local nash equilibrium.
\newblock In \emph{NeurIPS}, 2017.

\bibitem[Ho(2022)]{Ho2022ClassifierFreeDG}
Jonathan Ho.
\newblock Classifier-free diffusion guidance.
\newblock \emph{ArXiv}, abs/2207.12598, 2022.

\bibitem[Hong et~al.(2022)Hong, Tang, Li, Xiao, Wang, Li, Gu, Hu, Wang, Zhu, et~al.]{hong2022cogvideo}
Wenyi Hong, Ming Tang, Hongliang Li, Yongdong Xiao, Dongming Wang, Jingyuan Li, Xin Gu, Jiayi Hu, Haozhe Wang, Yuxiao Zhu, et~al.
\newblock Cogvideo: Large-scale pretraining for text-to-video generation via transformers.
\newblock \emph{arXiv preprint arXiv:2205.15868}, 2022.

\bibitem[Huang et~al.(2024)Huang, Huang, Liu, Yan, Lv, Liu, Xiong, Zhang, Chen, and Cao]{huang2024diffusion}
Yi Huang, Jiancheng Huang, Yifan Liu, Mingfu Yan, Jiaxi Lv, Jianzhuang Liu, Wei Xiong, He Zhang, Shifeng Chen, and Liangliang Cao.
\newblock Diffusion model-based image editing: A survey.
\newblock \emph{arXiv preprint arXiv:2402.17525}, 2024.

\bibitem[Jiang et~al.(2023{\natexlab{a}})Jiang, Yin, Xu, Jiang, Ouyang, Xiao, Qin, Xu, Qie, and Shen]{ccedit}
Ruoyu Jiang, Haoxin Yin, Zhongang Xu, Menghan Jiang, Yue Ouyang, Chao Xiao, Xiaogang Qin, Xiangyu Xu, Xiaohua Qie, and Chunhua Shen.
\newblock Ccedit: Creative and controllable video editing via diffusion models.
\newblock \emph{arXiv preprint arXiv:2309.16496}, 2023{\natexlab{a}}.

\bibitem[Jiang et~al.(2023{\natexlab{b}})Jiang, Gu, Jiang, Zheng, Yin, Loy, Qiao, and Guo]{fate-zero}
Yue Jiang, Chenyang Gu, Xiaoyu Jiang, Yiming Zheng, Hang Yin, Chen~Change Loy, Yu Qiao, and Ziwei Guo.
\newblock Fate-zero: Finetuning-free adaptation from text to edited video.
\newblock \emph{arXiv preprint arXiv:2312.03140}, 2023{\natexlab{b}}.

\bibitem[Karras et~al.(2023)Karras, Peng, Nurmi, Dai, Aittala, Lehtinen, and Aila]{dreampose}
Tero Karras, Shunsuke Peng, Aliaksandr Nurmi, Timo Dai, Miika Aittala, Jaakko Lehtinen, and Timo Aila.
\newblock Dreampose: Fashion video synthesis with stable diffusion.
\newblock \emph{arXiv preprint arXiv:2304.06025}, 2023.

\bibitem[Kim et~al.(2023)Kim, Jang, and Ye]{ground-a-video}
Hyeonho Kim, Gunhee Jang, and Jong~Chul Ye.
\newblock Ground-a-video: Zero-shot grounded video editing using text-to-image diffusion models.
\newblock \emph{arXiv preprint arXiv:2310.01107}, 2023.

\bibitem[Ku et~al.(2024)Ku, Wei, Ren, Yang, and Chen]{anyv2v}
Max Ku, Cong Wei, Weiming Ren, Huan Yang, and Wenhu Chen.
\newblock Anyv2v: A plug-and-play framework for any video-to-video editing tasks.
\newblock \emph{arXiv e-prints}, pages arXiv--2403, 2024.

\bibitem[Lab(2024)]{hunyuanvideo2024}
Tencent~AI Lab.
\newblock Hunyuan video: Large-scale text-to-video generation model.
\newblock \url{https://hunyuan.tencent.com/}, 2024.
\newblock Accessed: 2024-05-01.

\bibitem[Labs(2024)]{pikalabs}
Pika Labs.
\newblock Pika labs: Ai video generation.
\newblock \url{https://pika.art/}, 2024.
\newblock Accessed: 2024-05-01.

\bibitem[Leake et~al.(2017)Leake, Davis, Truong, and Agrawala]{leake2017computational}
Mackenzie Leake, Abe Davis, Anh Truong, and Maneesh Agrawala.
\newblock Computational video editing for dialogue-driven scenes.
\newblock In \emph{ACM Transactions on Graphics (TOG)}, pages 1--14, 2017.

\bibitem[Liu et~al.(2023)Liu, Zeng, Ren, Li, Zhang, Yang, Li, Yang, Su, Zhu, et~al.]{liu2023grounding}
Shilong Liu, Zhaoyang Zeng, Tianhe Ren, Feng Li, Hao Zhang, Jie Yang, Chunyuan Li, Jianwei Yang, Hang Su, Jun Zhu, et~al.
\newblock Grounding dino: Marrying dino with grounded pre-training for open-set object detection.
\newblock \emph{arXiv preprint arXiv:2303.05499}, 2023.

\bibitem[Liu et~al.(2024)Liu, Zhu, Shi, Zhang, Lou, Yang, Xi, Cao, Gu, Li, Li, Fang, Chen, Hsieh, Huang, Cheng, Nath, Hu, Liu, Krishna, Xu, Wang, Molchanov, Kautz, Yin, Han, and Lu]{liu2024nvila}
Zhijian Liu, Ligeng Zhu, Baifeng Shi, Zhuoyang Zhang, Yuming Lou, Shang Yang, Haocheng Xi, Shiyi Cao, Yuxian Gu, Dacheng Li, Xiuyu Li, Yunhao Fang, Yukang Chen, Cheng-Yu Hsieh, De-An Huang, An-Chieh Cheng, Vishwesh Nath, Jinyi Hu, Sifei Liu, Ranjay Krishna, Daguang Xu, Xiaolong Wang, Pavlo Molchanov, Jan Kautz, Hongxu Yin, Song Han, and Yao Lu.
\newblock Nvila: Efficient frontier visual language models, 2024.

\bibitem[Loshchilov and Hutter(2017)]{Loshchilov2017DecoupledWD}
Ilya Loshchilov and Frank Hutter.
\newblock Decoupled weight decay regularization.
\newblock In \emph{International Conference on Learning Representations}, 2017.

\bibitem[Ni et~al.(2023)Ni, Cao, Xu, Yin, and Tong]{flowvid}
Feng Ni, Yang Cao, Lan Xu, Mingyu Yin, and Xin Tong.
\newblock Flowvid: Taming imperfect optical flows for consistent video-to-video synthesis.
\newblock \emph{arXiv preprint arXiv:2312.17681}, 2023.

\bibitem[Oquab et~al.(2023)Oquab, Darcet, Moutakanni, Vo, Szafraniec, Khalidov, Fernandez, Haziza, Massa, El-Nouby, Assran, Ballas, Galuba, Howes, Huang, Li, Misra, Rabbat, Sharma, Synnaeve, Xu, J{\'e}gou, Mairal, Labatut, Joulin, and Bojanowski]{Oquab2023DINOv2LR}
Maxime Oquab, Timoth{\'e}e Darcet, Th{\'e}o Moutakanni, Huy~Q. Vo, Marc Szafraniec, Vasil Khalidov, Pierre Fernandez, Daniel Haziza, Francisco Massa, Alaaeldin El-Nouby, Mahmoud Assran, Nicolas Ballas, Wojciech Galuba, Russ Howes, Po-Yao~(Bernie) Huang, Shang-Wen Li, Ishan Misra, Michael~G. Rabbat, Vasu Sharma, Gabriel Synnaeve, Huijiao Xu, Herv{\'e} J{\'e}gou, Julien Mairal, Patrick Labatut, Armand Joulin, and Piotr Bojanowski.
\newblock Dinov2: Learning robust visual features without supervision.
\newblock \emph{ArXiv}, abs/2304.07193, 2023.

\bibitem[Ouyang et~al.(2023)Ouyang, Jiang, Yin, Xu, Xiao, Qin, Xu, Qie, and Shen]{dragvideo}
Yue Ouyang, Menghan Jiang, Haoxin Yin, Zhongang Xu, Chao Xiao, Xiaogang Qin, Xiangyu Xu, Xiaohua Qie, and Chunhua Shen.
\newblock Dragvideo: Interactive drag-style video editing.
\newblock \emph{arXiv preprint arXiv:2312.02216}, 2023.

\bibitem[Radford et~al.(2021)Radford, Kim, Hallacy, Ramesh, Goh, Agarwal, Sastry, Askell, Mishkin, Clark, et~al.]{radford2021learning}
Alec Radford, Jong~Wook Kim, Chris Hallacy, Aditya Ramesh, Gabriel Goh, Sandhini Agarwal, Girish Sastry, Amanda Askell, Pamela Mishkin, Jack Clark, et~al.
\newblock Learning transferable visual models from natural language supervision.
\newblock In \emph{International Conference on Machine Learning}, pages 8748--8763. PMLR, 2021.

\bibitem[Rana et~al.(2021)Rana, Singh, Valenzise, Dufaux, Komodakis, and Smolic]{rana2021deepremaster}
Abhijith Rana, Praveen Singh, Giuseppe Valenzise, Frederic Dufaux, Nikos Komodakis, and Aljosa Smolic.
\newblock Deepremaster: Temporal source-reference attention networks for comprehensive video enhancement.
\newblock \emph{ACM Transactions on Graphics}, 40\penalty0 (4):\penalty0 1--13, 2021.

\bibitem[Rasley et~al.(2020)Rasley, Rajbhandari, Ruwase, and He]{Rasley2020DeepSpeedSO}
Jeff Rasley, Samyam Rajbhandari, Olatunji Ruwase, and Yuxiong He.
\newblock Deepspeed: System optimizations enable training deep learning models with over 100 billion parameters.
\newblock \emph{Proceedings of the 26th ACM SIGKDD International Conference on Knowledge Discovery \& Data Mining}, 2020.

\bibitem[Ravi et~al.(2023)Ravi, Sharma, Mirzaei, Tao, Madhava, and Marculescu]{ravi2023sam2}
Nikhil Ravi, Abhishek Sharma, Sasha Mirzaei, Yin Tao, Krishna Madhava, and Radu Marculescu.
\newblock Sam2: Segment anything model at 15 hz with enhanced efficiency and semantic awareness.
\newblock \emph{arXiv preprint arXiv:2312.02245}, 2023.

\bibitem[Ren et~al.(2023)Ren, Xie, Chai, Zeng, Yuan, Zhao, and Zhao]{follow-your-pose}
Yue Ren, Yutong Xie, Donghao Chai, Zheng Zeng, Zhangyang Yuan, Qian Zhao, and Zhou Zhao.
\newblock Follow your pose: Pose-guided text-to-video generation using pose-free videos.
\newblock \emph{arXiv preprint arXiv:2304.01186}, 2023.

\bibitem[Song et~al.(2020)Song, Meng, and Ermon]{Song2020DenoisingDI}
Jiaming Song, Chenlin Meng, and Stefano Ermon.
\newblock Denoising diffusion implicit models.
\newblock \emph{ArXiv}, abs/2010.02502, 2020.

\bibitem[Sun et~al.(2024)Sun, Tu, Liao, and Tao]{sun2024diffusion}
Wenhao Sun, Rong-Cheng Tu, Jingyi Liao, and Dacheng Tao.
\newblock Diffusion model-based video editing: A survey.
\newblock \emph{arXiv preprint arXiv:2407.07111}, 2024.

\bibitem[Touvron et~al.(2023)Touvron, Martin, Stone, Albert, Almahairi, Babaei, Bashlykov, Batra, Bhargava, Bhosale, et~al.]{touvron2023llama}
Hugo Touvron, Louis Martin, Kevin Stone, Peter Albert, Amjad Almahairi, Yasmine Babaei, Nikolay Bashlykov, Soumya Batra, Prajjwal Bhargava, Shruti Bhosale, et~al.
\newblock Llama 2: Open foundation and fine-tuned chat models.
\newblock \emph{arXiv preprint arXiv:2307.09288}, 2023.

\bibitem[Tu et~al.(2025)Tu, Luo, Chen, Ji, Bai, and Zhao]{tu2025videoanydoor}
Yuanpeng Tu, Hao Luo, Xi Chen, Sihui Ji, Xiang Bai, and Hengshuang Zhao.
\newblock Videoanydoor: High-fidelity video object insertion with precise motion control.
\newblock \emph{arXiv preprint arXiv:2501.01427}, 2025.

\bibitem[Unterthiner et~al.(2019)Unterthiner, van Steenkiste, Kurach, Marinier, Michalski, and Gelly]{unterthiner2019accurate}
Thomas Unterthiner, Sjoerd van Steenkiste, Karol Kurach, Raphael Marinier, Marcin Michalski, and Sylvain Gelly.
\newblock Towards accurate generative models of video: A new metric \& challenges.
\newblock In \emph{ICLR}, 2019.

\bibitem[Wang et~al.(2024{\natexlab{a}})Wang, Yin, Ding, Guo, Guo, and Luo]{wang2024stepfun}
Wenhao Wang, Hongxing Yin, Menghan Ding, Bin Guo, Yong Guo, and Ping Luo.
\newblock Stepfun: A step function parameterization for boosting video generation.
\newblock \emph{arXiv preprint arXiv:2405.18335}, 2024{\natexlab{a}}.

\bibitem[Wang et~al.(2023)Wang, Yuan, Zhang, Chen, Wang, Zhang, Shen, Zhao, and Zhou]{wang2023videocomposer}
Xiang Wang, Hangjie Yuan, Shiwei Zhang, Dayou Chen, Jiuniu Wang, Yingya Zhang, Yujun Shen, Deli Zhao, and Jingren Zhou.
\newblock Videocomposer: Compositional video synthesis with motion controllability.
\newblock In \emph{Advances in Neural Information Processing Systems}, 2023.

\bibitem[Wang et~al.(2024{\natexlab{b}})Wang, Bao, Weng, Feng, Yin, Yang, Zhang, Dai, Zhao, Wang, et~al.]{wang2024microcinema}
Yanhui Wang, Jianmin Bao, Wenming Weng, Ruoyu Feng, Dacheng Yin, Tao Yang, Jingxu Zhang, Qi Dai, Zhiyuan Zhao, Chunyu Wang, et~al.
\newblock Microcinema: A divide-and-conquer approach for text-to-video generation.
\newblock In \emph{Proceedings of the IEEE/CVF Conference on Computer Vision and Pattern Recognition}, pages 8414--8424, 2024{\natexlab{b}}.

\bibitem[Wu et~al.(2021)Wu, Huang, Zhang, Li, Ji, Yang, Sapiro, and Duan]{Wu2021GODIVAGO}
Chenfei Wu, Lun Huang, Qianxi Zhang, Binyang Li, Lei Ji, Fan Yang, Guillermo Sapiro, and Nan Duan.
\newblock Godiva: Generating open-domain videos from natural descriptions.
\newblock \emph{ArXiv}, abs/2104.14806, 2021.

\bibitem[Wu et~al.(2023)Wu, Xie, Jiang, Zhao, Niu, Pu, and Ding]{wu2023openvid}
Hongwei Wu, Jiahui Xie, Jing Jiang, Zhou Zhao, Zhiwu Niu, Jian Pu, and Junping Ding.
\newblock Openvid: A large-scale open-domain video dataset for vision-language understanding.
\newblock In \emph{Proceedings of the IEEE/CVF International Conference on Computer Vision}, pages 16943--16953, 2023.

\bibitem[Xing et~al.(2023)Xing, Zhou, Zhang, Ren, Gu, Zuo, and Luo]{magic-edit}
Yue Xing, Kang Zhou, Yilin Zhang, Zhouhangbo Ren, Xiaodong Gu, Wangmeng Zuo, and Ping Luo.
\newblock Magicedit: High-fidelity and temporally coherent video editing.
\newblock \emph{arXiv preprint arXiv:2308.14749}, 2023.

\bibitem[Xu et~al.(2023)Xu, Shen, Zhou, Zhu, Zhuang, Loy, Dai, and Qiao]{magic-animate}
Zhongcong Xu, Jianfeng Shen, Xianzheng Zhou, Yiji Zhu, Xintao Zhuang, Chen~Change Loy, Bo Dai, and Yu Qiao.
\newblock Magicanimate: Temporally consistent human image animation using diffusion model.
\newblock \emph{arXiv preprint arXiv:2311.16498}, 2023.

\bibitem[Yang et~al.(2023)Yang, Gu, Zhang, Zhang, Chen, Sun, Chen, and Wen]{yang2023paint}
Binxin Yang, Shuyang Gu, Bo Zhang, Ting Zhang, Xuejin Chen, Xiaoyan Sun, Dong Chen, and Fang Wen.
\newblock Paint by example: Exemplar-based image editing with diffusion models.
\newblock In \emph{Proceedings of the IEEE/CVF conference on computer vision and pattern recognition}, pages 18381--18391, 2023.

\bibitem[Yang et~al.(2024{\natexlab{a}})Yang, Ding, Zhu, Xiao, Liu, Wang, Feng, Xia, Guo, Gao, et~al.]{sora2024}
Yingqing Yang, Mengchen Ding, Wenjing Zhu, Longteng Xiao, Ruoyi Liu, Shilong Wang, Yunfei Feng, Tao Xia, Yong Guo, Mingming Gao, et~al.
\newblock Sora: A review on background, technology, limitations, and opportunities of large vision models.
\newblock \emph{arXiv preprint arXiv:2402.17177}, 2024{\natexlab{a}}.

\bibitem[Yang et~al.(2024{\natexlab{b}})Yang, Teng, Zheng, Ding, Huang, Xu, Yang, Hong, Zhang, Feng, et~al.]{yang2024cogvideox}
Zhuoyi Yang, Jiayan Teng, Wendi Zheng, Ming Ding, Shiyu Huang, Jiazheng Xu, Yuanming Yang, Wenyi Hong, Xiaohan Zhang, Guanyu Feng, et~al.
\newblock Cogvideox: Text-to-video diffusion models with an expert transformer.
\newblock \emph{arXiv preprint arXiv:2408.06072}, 2024{\natexlab{b}}.

\bibitem[Ye et~al.(2023)Ye, Zhang, Liu, Han, and Yang]{Ye2023IPAdapterTC}
Hu Ye, Jun Zhang, Siyi Liu, Xiao Han, and Wei Yang.
\newblock Ip-adapter: Text compatible image prompt adapter for text-to-image diffusion models.
\newblock \emph{ArXiv}, abs/2308.06721, 2023.

\bibitem[Yin et~al.(2023)Yin, Jiang, Yan, Ouyang, Xu, Xiao, Qin, Xu, Qie, and Shen]{drag-a-video}
Hao Yin, Menghan Jiang, Zhaoyang Yan, Yue Ouyang, Zhongang Xu, Chao Xiao, Xiaogang Qin, Xiangyu Xu, Xiaohua Qie, and Chunhua Shen.
\newblock Drag-a-video: Non-rigid video editing with point-based interaction.
\newblock \emph{arXiv preprint arXiv:2312.02936}, 2023.

\bibitem[Yu et~al.(2023)Yu, Dai, Dong, Qiao, and Zhu]{yu2023deficiency}
Gangwei Yu, Xizhou Dai, Jianmin Dong, Yu Qiao, and Xiaogang Zhu.
\newblock Deficiency-aware masked video transformer for video inpainting.
\newblock \emph{arXiv preprint arXiv:2303.08467}, 2023.

\bibitem[Zhou et~al.(2023)Zhou, Yang, Zhang, Liao, Chen, Lu, Zhao, Liu, and Zhang]{zhou2023propainter}
Wenbo Zhou, Yujun Yang, Xiang Zhang, Hang Liao, Weiming Chen, Fang Lu, Shikui Zhao, Yingqing Liu, and Ziwei Zhang.
\newblock Propainter: Improving propagation and transformer for video inpainting.
\newblock In \emph{Proceedings of the IEEE/CVF International Conference on Computer Vision}, pages 21559--21568, 2023.

\end{thebibliography}
}

\end{document}